%% file: root.tex
\documentclass[letterpaper, 10 pt, conference]{ieeeconf}  
\IEEEoverridecommandlockouts                              
\input{latex_settings_tomoya-y}

\usepackage{cite}
\usepackage{comment}
\usepackage{color}
\usepackage{soul}
\usepackage[font=footnotesize]{subcaption}
\usepackage[font=footnotesize]{caption}

\title{\LARGE \bf
    Randomized-to-Canonical Model Predictive Control \\ for Real-world Visual Robotic Manipulation
}

\author{
    Tomoya Yamanokuchi$^{1}$, 
    Yuhwan Kwon$^{1}$, 
    Yoshihisa Tsurumine$^{1}$, \\ 
    Eiji Uchibe$^{2}$, 
    Jun Morimoto$^{3,2}$, 
    and Takamitsu Matsubara$^{1}$
    \thanks{
        This work was supported by JST-Mirai Program Grant Number JPMJMI18B8 and JPMJMI21B1, Japan.
    }
    \thanks{
        $^{1}$T. Yamanokuchi, Y. Kwon, Y. Tsurumine, and T. Matsubara are with the Graduate School of Information Science, Nara Institute of Science and Technology (NAIST), Nara, Japan.
    }
    \thanks{
        $^{2}$E. Uchibe and J. Morimoto are the Advanced Telecommunications Research Institute International (ATR), Kyoto, Japan.
    }
    \thanks{
        $^{3}$J. Morimoto is with the Department of Systems Science, Graduate School of Informatics, Kyoto University, Kyoto, Japan.
    }
}

\begin{document}

\maketitle

\newcommand{\figcaption}[1]{\def\@captype{figure}\caption{#1}}
\newcommand{\tblcaption}[1]{\def\@captype{table}\caption{#1}}
\makeatother

\begin{abstract}
    Many works have recently explored Sim-to-real transferable visual model predictive control (MPC). 
    However, such works are limited to one-shot transfer, where real-world data must be collected once to perform the sim-to-real transfer, which remains a significant human effort in transferring the models learned in simulations to new domains in the real world. To alleviate this problem, we first propose a novel model-learning framework called Kalman Randomized-to-Canonical Model (KRC-model). This framework is capable of extracting task-relevant intrinsic features and their dynamics from randomized images. We then propose Kalman Randomized-to-Canonical Model Predictive Control (KRC-MPC) as a zero-shot sim-to-real transferable visual MPC using KRC-model. The effectiveness of our method is evaluated through a valve rotation task by a robot hand in both simulation and the real world, and a block mating task in simulation. The experimental results show that KRC-MPC can be applied to various real domains and tasks in a zero-shot manner. 
\end{abstract}

\section{Introduction}
    Model predictive control (MPC) is widely used in robot control as an attractive method that is robust to modeling errors and can be applied to various tasks by adjusting the cost function.
    With the development of deep learning, visual MPC, which performs MPC using images, has achieved remarkable results \cite{Finn_ICRA_2017, Yen-Chen_CoRL_2019, Limoyo_IEEE_Robotics_2020} in recent years.
    However, the data-collection cost in the real world is a major common problem for visual MPC, since learning visual dynamics models requires many image data. 
    To overcome this problem, sim-to-real transfer approaches have been explored \cite{Ryan_RSS_2020,Suraj_ICRA_2020}; however, they have a limitation as described below.
    
    The limitation is that the `goal image' must be provided for each test domain before running MPC. In previous studies \cite{Ryan_RSS_2020,Suraj_ICRA_2020}, the dynamics was modeled as a time evolution from random images to random images, resulting in a domain adaptive model. This need for such a one-shot (domain-adaptation) procedure may seriously limit the method's applications. Suraj et al. \cite{Suraj_ICRA_2020} tackled this problem by learning an additional network that generates domain-dependent goal images; however, it cannot be transferred to different tasks.  
    
    Our idea for alleviating this limitation is to extract task-relevant {\it intrinsic features} and their dynamics from randomized images. In the context of static image transformation, James et al. \cite{James_CVPR_2019} showed that by introducing {\it canonical images}, in which each task-relevant object is visually identifiable, domain-dependent and task-irrelevant information in object color, lighting conditions, background, etc. can be removed from the input images, and intrinsic (geometric) features useful for subsequent manipulation tasks can be extracted. Given such 
    task-relevant features and their dynamics guided by canonical images, it may be possible to perform zero-shot visual MPC on unknown test domains without any test data for domain adaptation. 

\begin{figure}[t]
\centering
    \includegraphics[width=\columnwidth]{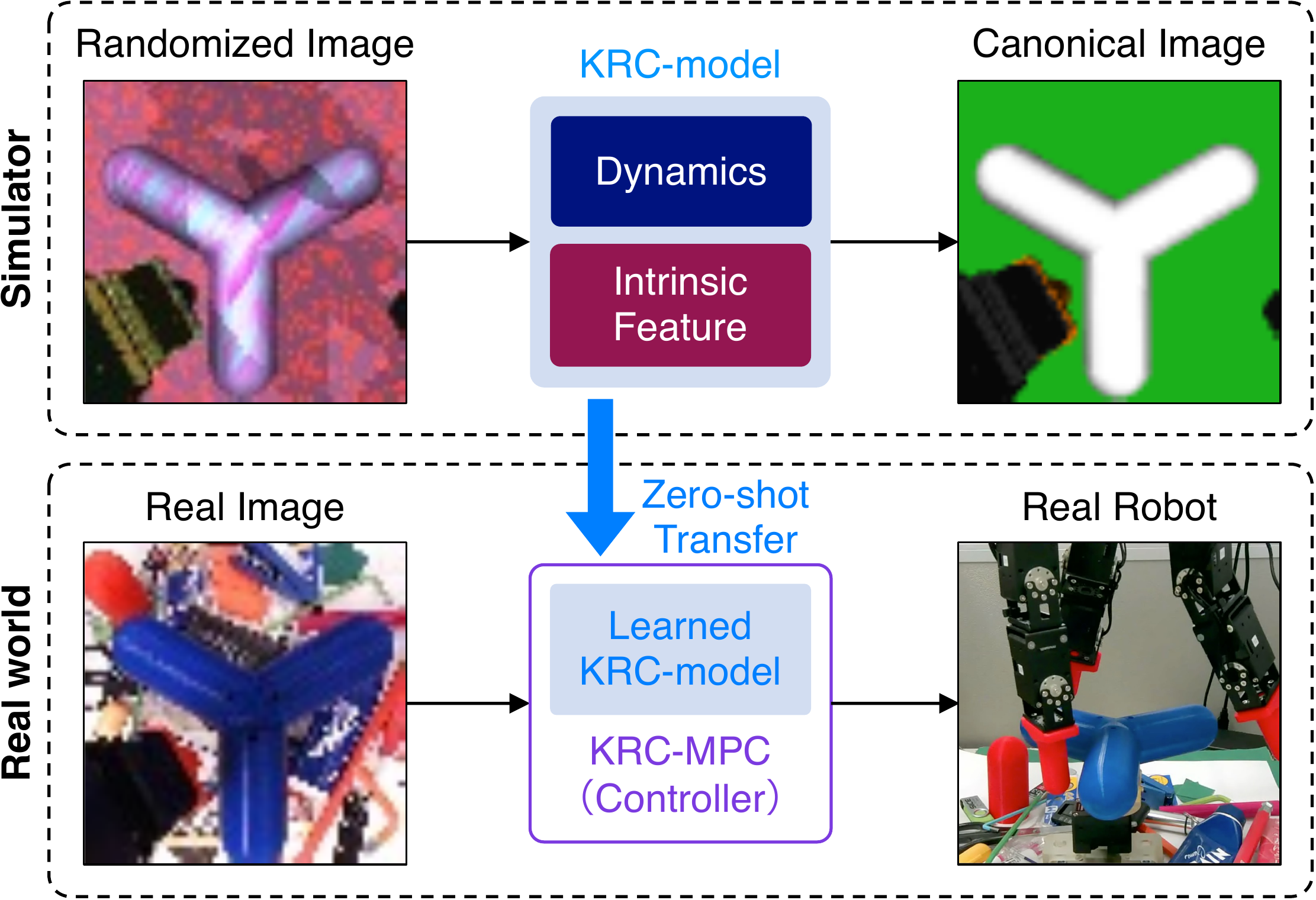}
    \caption{
    Overview of proposed method. 
    KRC-model jointly learns task-relevant intrinsic features and nonlinear dynamics via randomized-to-canonical structure. 
    KRC-MPC with KRC-model can be transferred to the real world in a zero-shot manner.
    In this study, we evaluated our method through a valve rotation task and a block mating task, where the task-relevant features are geometric features of the objects and robot.
    Thus, in the case of the valve rotation task, canonical images are defined as images in which the valve and robot are identified by their colors (white valve, black robot finger bodies, orange robot fingertips).
    }
    \label{fig:overview_of_mehod}
\end{figure}

    In this paper, we first propose a novel sim-to-real transferable visual dynamics model called \textit{Kalman Randomized-to-Canonical Model (KRC-model)}, which can extract the intrinsic features and their dynamics from the randomized images by exploiting the key property: randomized-to-canonical feature extraction. Specifically, KRC-model integrates the following three components: 
    Randomized-to-Canonical Adaptation Network \cite{James_CVPR_2019}, 
    Linear Gaussian State Space Model \cite{LGSSM}, and 
    Dynamics Parameter Network \cite{KVAE}. 
    A variational learning algorithm exploiting the merits of the structure for learning all of the models in a unified and computationally efficient framework is also presented. We then propose Kalman Randomized-to-Canonical MPC (KRC-MPC) as a zero-shot sim-to-real transferable visual MPC using KRC-model, requiring no real-world data for either model learning or visual MPC. 
    Our KRC-MPC is the first visual MPC framework that is zero-shot sim-to-real transferable beyond the reality gap in appearance. 
    An overview of our method is shown in Fig.~\ref{fig:overview_of_mehod}. 

    We applied our method to a valve rotation task with a robot hand in both simulation and the real world, and a block mating task in simulation to evaluate its effectiveness in various appearances of domains and various tasks.
    The effect of canonical image is also evaluated through ablation studies by comparing the cases using inappropriate canonical images.
    The experimental results show that KRC-MPC with KRC-model can be transferred to various appearances of domains in a zero-shot manner.

    Our contributions are as follows:
    \begin{itemize}
        \item We propose a framework for zero-shot sim-to-real transferable visual MPC (KRC-MPC).
        \item We propose a visual dynamics model to achieve KRC-MPC (KRC-model).
        \item We evaluate the effectiveness of our method in both simulation and the real world.
    \end{itemize}

\section{Related works}\label{Related_work}
    The capabilities of visual MPC for visual robotic manipulation has been demonstrated in many studies\cite{Finn_ICRA_2017,Yen-Chen_CoRL_2019,Limoyo_IEEE_Robotics_2020}.
    Finn et al.\cite{Finn_ICRA_2017} successfully performed pushing tasks with visual MPC by learning a visual dynamics model from unlabeled data collected by ten robot arms in the real world.
    Yen-Chen et al.\cite{Yen-Chen_CoRL_2019} succeeded in pushing for unknown objects with visual MPC by jointly learning a visual dynamics model and context embedding from experience data.
    Limoyo et al.\cite{Limoyo_IEEE_Robotics_2020} incorporated a novelty-detection mechanism in the visual dynamics model and successfully performed reaching tasks in uncertain environments.
    Furthermore, a few recent studies have examined approaches of transfer learning and generalization for visual MPCs \cite{Sudeep_CoRL2019, Edward_ICLR2022}. However, the cost of collecting real-world data is still high and a common challenge.

    A promising approach to solving this problem is sim-to-real transfer, and many techniques have been proposed, mainly in the context of policy transfer for model-free reinforcement learning \cite{Sadeghi_RSS_2017,Pinto_RSS_2018,James_CVPR_2019,Andrychowicz_IJRR_2020,James_CoRL_2017}, or imitation learning \cite{Alessandro_RAL_2020}.
    These methods randomize sensor information, which serves as input for the policy, to obtain generalization performance and enable transfer to the real-world environments. However, such policies, which typically focus on a single task, are more difficult to transfer to other tasks than the model-based methods.
    Recently, several studies\cite{Ryan_RSS_2020,Suraj_ICRA_2020} have been conducted on sim-to-real transferable visual MPC.
    Ryan et al.\cite{Ryan_RSS_2020} successfully transferred visual MPC for multi-task cloth manipulation to the real world by applying visual randomization to train a visual dynamics model.
    Suraj et al.\cite{Suraj_ICRA_2020} also used a similar approach to transfer visual MPC in the block manipulation task to the real world.
    However, in these studies, a goal image needs to be demonstrated in advance for each test domain to execute MPC because they model dynamics as a time evolution from random images to random images, which leads to the domain-adaptive model. 
    While Suraj et al. \cite{Suraj_ICRA_2020} tackled this problem by learning an additional network that generates domain-dependent goal images, the generator needs to be trained for each task.
    
    In contrast to those studies, we propose a framework of visual MPC that can be transferred to the real world in a zero-shot manner based on the visual dynamics model guided by canonical images.

\begin{figure}[t]
\centering%
  \begin{tikzpicture}[bend angle=30,>=latex,font=\normalsize,scale=0.65, every node/.style={transform shape}]%
	\tikzstyle{obs} = [ circle,  thick,  draw = black!100, fill = blue!10,  minimum size = 1.0cm, inner sep = 0pt]
	\tikzstyle{lat} = [ circle,  thick,  draw = black!100, fill = red!0,    minimum size = 1.0cm, inner sep = 0pt]
	\tikzstyle{par} = [ circle,  thin,   draw, fill = black!100,            minimum size = 0.8cm, inner sep = 0pt]
	\tikzstyle{det} = [ diamond, thick,  draw = black!100, fill = red!0,    minimum size = 1.25cm, inner sep = 0pt]
	\tikzstyle{inv} = [ circle,  thin,   draw=white!100, fill = white!100,  minimum size = 0.8cm, inner sep = 0pt]
	\tikzstyle{annotation} = [rectangle, thin, draw=white, fill=white ]
	\tikzstyle{every label} = [black!100]%

	\begin{scope}[node distance = 1.7cm and 1.7cm, rounded corners=0pt]
	    \node       (z_tm2) [] {};
		\node [lat] (z_tm1) [ right =0.7cm of z_tm2] {$\zb_{t-1}$};
		\node [lat] (z_t)   [ right =3.35cm of z_tm1] {$\zb_{t}$};
		\node [lat] (z_tp1) [ right =3.35cm of z_t]   {$\zb_{t+1}$};
		\node       (z_tp2) [ right =2.4cm of z_tp1] {};
        \draw[post] (z_tm2) edge  (z_tm1);
		\draw[post] (z_tm1) edge  (z_t);
		\draw[post] (z_t) edge  (z_tp1);
		\draw[post] (z_tp1) edge  (z_tp2);
	    \node (a_tm2) [ above of = z_tm2] {};
		\node [lat] (a_tm1) [ above=1.1cm of z_tm1]   {$\ab_{t-1}$};
		\node [lat] (a_t)   [ above=1.1cm of z_t] {$\ab_{t}$};
		\node [lat] (a_tp1) [ above=1.1cm of z_tp1] {$\ab_{t+1}$};
		\node (a_tp2) [ right of = a_tp1]{};
		\draw[post] (z_tm1) edge  (a_tm1);
		\draw[post] (z_t) edge  (a_t);
		\draw[post] (z_tp1) edge  (a_tp1);
		\node [obs, color=black!13, text=black, draw=black] (a_aux_tm1) [ below right= 1.8cm and 1.45cm of z_tm1 ]   {$\bb_{t-1}$};
		\node [obs, color=black!13, text=black, draw=black] (a_aux_t)   [ below right= 1.8cm and 1.45cm of z_t ] {$\bb_{t}$};
		\node [obs, color=black!13, text=black, draw=black] (a_aux_tp1) [ below right= 1.8cm and 1.45cm of z_tp1 ] {$\bb_{t+1}$};
		\draw[post] (z_tm1) edge  (a_aux_tm1);
		\draw[post] (z_t) edge  (a_aux_t);
		\draw[post] (z_tp1) edge  (a_aux_tp1);
		\node [obs, color=black!13, text=black, draw=black] (x_tm1_can) [ above right=1.0cm and -0.1cm of a_tm1] {$\xb_{t-1}^{\textrm{can}}$};
		\node [obs, color=black!13, text=black, draw=black] (x_t_can)   [ above right=1.0cm and -0.1cm of a_t]   {$\xb_{t}^{\textrm{can}}$};
		\node [obs, color=black!13, text=black, draw=black] (x_tp1_can) [ above right=1.0cm and -0.1cm of a_tp1] {$\xb_{t+1}^{\textrm{can}}$};
		\node [obs, color=black!13, text=black, draw=black] (x_tm1_ran) [ above left=1.0cm and -0.1cm of a_tm1]   {$\xb_{t-1}^{\textrm{ran}}$};
		\node [obs, color=black!13, text=black, draw=black] (x_t_ran)   [ above left=1.0cm and -0.1cm of a_t] {$\xb_{t}^{\textrm{ran}}$};
		\node [obs, color=black!13, text=black, draw=black] (x_tp1_ran) [ above left=1.0cm and -0.1cm of a_tp1] {$\xb_{t+1}^{\textrm{ran}}$};
	    \node (d_tm2) [below left= 0.42cm and 0.8cm of z_tm1] {};
		\node [det] (d_tm1) [color=DPNColor, fill=white, below right= 0.22cm and 1.45cm of z_tm1]   {$\db_{t}$};
		\node [det] (d_t)   [color=DPNColor, fill=white, below right= 0.22cm and 1.45cm of z_t]   {$\db_{t+1}$};
		\node [det] (d_tp1) [color=DPNColor, fill=white, below right= 0.22cm and 1.45cm of z_tp1] {$\db_{t+2}$};
        \node (DPNColor,d_tp2) [ right of = d_tp1]{};
        \draw[DPNColor,post] (d_tm2) edge  (d_tm1);
		\draw[DPNColor, post] (d_tm1) edge  (d_t);
		\draw[DPNColor, post] (d_t) edge  (d_tp1);

        \node (alpha_tm2) [above left= 0.5cm and 0.85cm of z_tm1] {};
		\node [det] (alpha_tm1) [color=DPNColor, fill=white, above right= 0.42cm and 1.45cm of z_tm1]   {$\alphab_{t}$};
		\node [det] (alpha_t)   [color=DPNColor, fill=white, above right= 0.42cm and 1.45cm of z_t]   {$\alphab_{t+1}$};
		\node [det] (alpha_tp1) [color=DPNColor, fill=white, above right= 0.42cm and 1.45cm of z_tp1] {$\alphab_{t+2}$};
        \draw[DPNColor, post]  (d_tm1) edge  (alpha_tm1);
		\draw[DPNColor, post] (d_t) edge  (alpha_t);
		\draw[DPNColor, post] (d_tp1)   edge  (alpha_tp1);

        \draw[DPNColor,post] (alpha_tm2) edge  (a_tm1);
        \draw[DPNColor,post] (alpha_tm2) edge  (z_tm1);
        
		\draw[DPNColor,post] (alpha_tm1) edge  (z_t);
		\draw[DPNColor,post] (alpha_t) edge  (z_tp1);
		\draw[DPNColor,post] (alpha_tm1) edge  (a_t);
		\draw[DPNColor,post] (alpha_t) edge  (a_tp1);

		\draw[DPNColor,post] (a_aux_tm1) edge  (d_tm1);
		\draw[DPNColor,post] (a_aux_t) edge  (d_t);
		\draw[DPNColor,post] (a_aux_tp1) edge  (d_tp1);
		\draw[DPNColor,post] (a_tm1) edge  (d_tm1);
		\draw[DPNColor,post] (a_t) edge  (d_t);
		\draw[DPNColor,post] (a_tp1) edge  (d_tp1);
		\node [obs, color=black!13, text=black, draw=black] (u_tm1) [ below=1.5cm of z_tm1]   {$\ub_{t-1}$};
		\node [obs, color=black!13, text=black, draw=black] (u_t) [ below=1.5cm of  z_t] {$\ub_{t}$};
		\node [obs, color=black!13, text=black, draw=black] (u_tp1) [ below=1.5cm of z_tp1] {$\ub_{t+1}$};
		\draw[post] (u_tm1) edge  (z_tm1);
		\draw[post] (u_t) edge  (z_t);
		\draw[post] (u_tp1) edge  (z_tp1);
		\draw[DPNColor,post] (u_tm1) edge  (d_tm2);
		\draw[DPNColor,post] (u_t) edge  (d_tm1);
		\draw[DPNColor,post] (u_tp1) edge  (d_t);
		\draw[post] (x_tp1_ran) edge[bend right, dashed]  (a_tp1);
		\draw[post] (x_t_ran)   edge[bend right, dashed]  (a_t);
		\draw[post] (x_tm1_ran) edge[bend right, dashed]  (a_tm1);
		
		\draw[post] (a_tp1) edge[bend right]  (x_tp1_can);
		\draw[post] (a_t)   edge[bend right]  (x_t_can);
		\draw[post] (a_tm1) edge[bend right]  (x_tm1_can);
		
		\node[annotation, text=RCANColor] (vae) at (11.45cm, 2.4cm) {RCAN};
		\node[annotation, text=Cerulean] (lgssm) at (6.8cm, -3.4cm) {LGSSM};
		\node[annotation, text=DPNColor] (dpn) at (9.3cm, -1.3cm) {DPN};
		\begin{pgfonlayer}{background}
	    \filldraw [dotted, line width = 1.2pt, draw=Cerulean, fill=white]
		($(a_tm1.north west) + (-1cm, 0.5cm)$) rectangle ($(u_tp1.south east) + (2.65cm, -0.5cm)$);
		\filldraw [dotted, line width = 1.2pt, draw=RCANColor, fill=white]
		($(x_tp1_can.north west) + (-1.6cm, 0.4cm)$) rectangle ($(a_tp1.south east) + (1.0cm, -0.5cm)$);
     	\end{pgfonlayer}
	\end{scope}%
\end{tikzpicture}%
\caption{
    Graphical model of KRC-model, which consists of RCAN (red dotted rectangle), LGSSM (blue dotted rectangle), and DPN (orange path). Shaded nodes denote observed variables, while non-shaded nodes denote latent variables. The solid arrow represents the generative model, and the dashed arrow represents the inference model.
}
\label{fig:KRCAN}%
\end{figure}
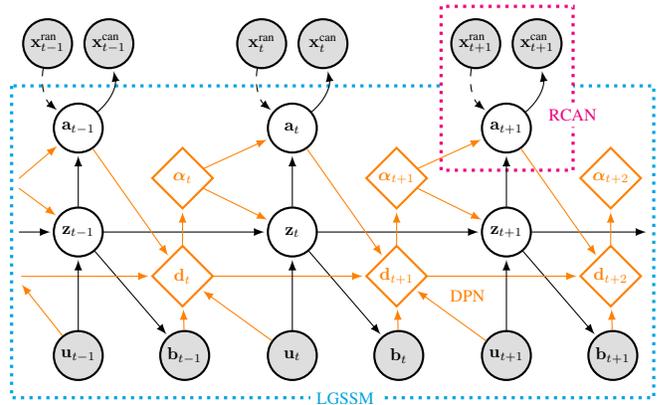%

\section{Proposed method} \label{Proposed_method}
    Our goal of model learning is to extract the intrinsic features and their dynamics from randomized images. To that end, in this section, we first describe our visual dynamics model, KRC-model, and then derive its efficient model-learning algorithm based on the variational inference. 
    We then describe KRC-MPC, a visual MPC that uses KRC-model.

    \subsection{Visual Dynamics Model}
        KRC-model consists of Randomized-to-Canonical Adaptation Network (RCAN) \cite{James_CVPR_2019}, Linear Gaussian State Space Model (LGSSM) \cite{LGSSM}, and Dynamics Parameter Network (DPN) \cite{KVAE}. RCAN is used to extract the intrinsic features from randomized images with the aid of canonical images. Then, the associated dynamics with the intrinsic features is captured through LGSSM. DPN determines the time-varying matrix parameters of LGSSM.
        Fig.~\ref{fig:KRCAN} shows a graphical model of KRC-model.

    \subsubsection{RCAN}
        We model RCAN as a feature extractor with an encoder-decoder structure as follows:
        \begin{numcases}{}
            \ab_t                & $\sim q_{\phi}(\ab_t|\xb^\textrm{ran}_t)$, \\
            \xb^\textrm{can}_t   & $\sim p_{\theta}(\xb^\textrm{can}_t|\ab_t)$, 
        \end{numcases}
        where $q_{\phi}$ is an encoder with parameter $\phi$, $p_{\theta}$ is a decoder with parameter $\theta$, and $\xb^\textrm{can}_t \in \Rbb^{d_W \times d_V}, \xb^\textrm{ran}_t \in \Rbb^{d_W \times d_V}, \ab_t \in \Rbb^{d_\ab}$ are canonical image, randomized image, and intrinsic feature at time step $t$, respectively.

    \subsubsection{LGSSM} \label{LGSSM}
        We use a time-varying LGSSM as a dynamics model, inspired by KVAE \cite{KVAE}.
        Given the observation as $\yb_t \in \Rbb^{d_\yb}$, state as $\zb_t \in \Rbb^{d_\zb}$, and control input as $\ub_t \in \Rbb^{d_\ub}$, the state transition model and the observation model are defined as follows:
        \begin{align}
            p_{\gamma_t, \psi}({\bf z}_t \mid {\bf z}_{t-1}, {\bf u}_t) &= {\mathcal N}({\bf z}_t \mid {\bf A}_t {\bf z}_{t-1} + {\bf B}_t {\bf u}_t, {\bf Q}), \label{eq: LGSSM-transition} \\ 
            p_{\gamma_t, \psi}({\bf y}_t \mid {\bf z}_t) &= {\mathcal N}({\bf y}_t \mid {\bf C}_t {\bf z}_t, {\bf R}), \label{eq: LGSSM-observation} 
        \end{align}
        where $\yb_t = [\ab_t, \bb_t]$, $\bb_t \in \Rbb^{d_\bb}$ is the sensor information of the robot, and $\gamma_t = [{\bf A}_t, {\bf B}_t, {\bf C}_t]$ are the state transition matrix, control matrix, and observation matrix, respectively.
        Then, $\psi$ is the network that determines $\gamma_t$, and ${\bf Q} \in \Rbb^{d_\zb \times d_\zb}, {\bf R} \in \Rbb^{d_\yb \times d_\yb}$ are the covariance matrices of the process and observation noise.
        Assuming that the initial state is ${\bf z}_1 \sim {\mathcal N}({\bf z}_1 | {\bf 0}, {\bf \Sigma})$, the joint probability distribution of dynamics model can be expressed as follows:
        \begin{align}
            & p_{\gamma, \psi}({\bf y}, {\bf z} \mid {\bf u}) \notag \\
            & = \prod_{t=1}^T p_{{\gamma}_t, \psi} ({\bf y}_t \mid {\bf z}_t) \cdot p({\bf z}_1) \prod_{t=2}^T  p_{{\gamma}_t, \psi}({\bf z}_t \mid {\bf z}_{t-1}, {\bf u}_t), 
            \label{eq: LGSSM-join_prob} 
        \end{align}
        where $\yb=[\yb_1, \dots, \yb_T]$, $\zb=[\zb_1, \dots, \zb_T]$, $\ub=[\ub_1, \dots, \ub_T]$, and $\gamma=[\gamma_1, \dots, \gamma_T]$. From the perspective of dynamics learning, the intrinsic features $\ab_t$ allows the dynamics learning to be disentangled from the images, making for a tractable computation (Appendix).

    \subsubsection{Dynamics Parameter Network} \label{teian-modelka-DPN}
        We introduce DPN to determine the time-varying matrix parameter $\gamma_t$ of the dynamics model.
        DPN is modeled as a deterministic network that takes past observations $\yb_{0:t-1}$ and control inputs $\ub_{1:t}$ at each time step and outputs the weights $\alphab_t$ to determine $\gamma_t$ as shown in Eq.~\eqref{eq:DPN_definition}:			
        \begin{align} \label{eq:DPN_definition}
            \alphab_{t} = \psi(\yb_{0:t-1}, \ub_{1:t}).
        \end{align}
        The output $\alphab_t = [ \alpha_t^{(1)}, .., \alpha_t^{(K)}]$ of DPN is a $K$-dimensional vector that satisfies $\sum_{k=1}^K \alpha_t^{(k)} = 1$ and is used to make mixtures of $K$ different dynamics models.
        Therefore, each parameter of $\gamma_t = [\Ab_t, \Bb_t, \Cb_t]$ can be expressed as follows:
        \begin{align} \label{eq:drkvae-dpn-gamma-def}
        \gamma_t &= \left[ \sum_{k=1}^K \alpha_t^{(k)}\Ab^{(k)}, \sum_{k=1}^K \alpha_t^{(k)}\Bb^{(k)}, \sum_{k=1}^K \alpha_t^{(k)}\Cb^{(k)} \right].
        \end{align}
        The $K$ basis matrices $\Ab^{(k)}, \Bb^{(k)}, \Cb^{(k)}$ are trained globally over the entire dataset.

    \subsection{Model Learning Algorithms}
        We first derive a model learning algorithm in a general setting where the state variables are unknown and unobserved. Next, we present a more efficient variant that takes advantage of the fact that state variables are commonly known and available in sim-to-real setups.

    \subsubsection{With Unobserved States} \label{teian-modelgakusyu}
        First, we derive a model learning algorithm in a general setting where the state variables are unknown and unobserved.
        In such cases, the states $\zb$ are treated as latent variables.
        Therefore, we are interested in the posterior distribution of $\ab$ and $\zb$. 
        However, since this posterior cannot be computed analytically, we alternatively maximize the evidence lower bound (ELBO) of marginal log-likelihood $\Lcal = \sum_n^N \log p(\xb^{\textrm{can} (n)}, \bb^{(n)}| \ub^{(n)})$.
        For simplicity, we omit the sequence index $n$ in the following description.
        Since the joint probability distribution of KRC-model is expressed as
        \begin{align}
        p(\xb^{\textrm{can}}, \yb, \zb |\ub) = p_\theta(\xb^{\textrm{can}} | \ab) \, p_{\gamma, \psi}(\yb | \zb) \, p_{\gamma, \psi}(\zb | \ub),
        \label{eq: DRKVAE_joint_prob}%
        \end{align}
        ELBO can be written as
        \begin{align} \label{eq:KRCAN_ELBO_unclear}%
        & \log p(\xb^{\textrm{can}}, \bb | \ub) \\
        &= \log \int p(\xb^{\textrm{can}}, \ab, \bb, \zb | \ub) \drm \ab \drm \zb \\
        & \geq \left \langle \log \frac{p_{\theta}(\xb^{\textrm{can}}|\ab)p_{\gamma, \psi}(\yb|\zb)p_{\gamma, \psi}(\zb|\ub)}{q(\ab, \zb|\cdot)} \right \rangle_{q(\ab, \zb|\cdot)}\\
        & = \Fcal(\theta, \gamma, \psi, \phi),
        \end{align}
        where $q(\ab, \zb|\cdot)$ is the variational distribution.
        Here, recalling that the Kalman Smoother can analytically obtain the posterior distribution $p_{\gamma, \psi}(\zb | \yb, \ub)$ with LGSSM\cite{LGSSM}, and by utilizing the encoder $q_{\phi}(\ab|\xb^\textrm{ran})$, we can define the variational distribution as
        \begin{align}
        q(\ab, \zb |\xb^\textrm{ran}, \bb, \ub) = q_{\phi}(\ab|\xb^\textrm{ran}) p_{\gamma, \psi}(\zb | \yb, \ub).
        \end{align}
        Then, by using this variational distribution, ELBO becomes
        \begin{align} \label{eqn: elbo_2}
            & \Fcal(\theta, \gamma, \psi, \phi) = \bigg\langle \log \frac{p_{\theta}(\xb^\textrm{can} | \ab)}{q_{\phi}(\ab |\xb^{\textrm{ran}}) } \notag \\ 
            & + \left\langle \log \frac{p_{\gamma, \psi}(\yb | \zb) p_{\gamma, \psi}(\zb | \ub)}{p_{\gamma, \psi}(\zb | \yb, \ub)} \right\rangle_{p_{\gamma, \psi}(\zb | \yb, \ub)} \bigg\rangle_{q_{\phi}(\ab|\xb^\textrm{ran})}.
        \end{align}
        This ELBO value can be estimated by Monte Carlo integration using samples $\{ \widetilde{\ab}_i, \widetilde{\zb}_i\}_{i=1}^I$.

    \subsubsection{With Observed States} \label{teian-modelgakusyu_z_obs}
        Next, we present a more efficient variant that takes advantage of the fact that state variables are commonly known and available in sim-to-real setups.
        In such cases, we are only interested in the posterior distribution of $\ab$.
        However, since this posterior also cannot be computed analytically, we maximize ELBO of the marginal log-likelihood, as in Section \ref{teian-modelgakusyu_z_obs}.
        By introducing the encoder as a variational distribution, ELBO of the marginal log-likelihood $\Lcal = \sum_n^N \log p(\xb^{\textrm{can} (n)}, \bb^{(n)}, \zb^{(n)} | \ub^{(n)})$ becomes
        \begin{align} \label{eq:KRCAN_ELBO_clear}%
        \Fcal(\theta, \gamma, \psi, \phi) &= \left \langle \log \frac{p_{\theta}(\xb^{\textrm{can}}|\ab)}{q_{\phi}(\ab |\xb^{\textrm{ran}}) } + \log p_{\gamma, \psi}(\yb|\zb) \right \rangle_{q_{\phi}(\ab |\xb^\textrm{ran})} \notag \\
        & \hspace{5mm} + \log p_{\gamma, \psi}(\zb|\ub).
        \end{align}
        This ELBO can be estimated by Monte Carlo integration using samples $\{ \widetilde{\ab}_i\}_{i=1}^I$.
        
        The learning process of KRC-model with observed states is shown in Algorithm~\ref{alg:KRC-model}.
        We discussed the relationship between the dimension of the intrinsic feature $\ab$ and the computational cost for model learning in Appendix.

\begin{algorithm}[t]
  \SetKwData{Left}{left}\SetKwData{This}{this}\SetKwData{Up}{up}
  \SetKwFunction{Union}{Union}\SetKwFunction{FindCompress}{FindCompress}
  \SetKwInOut{Input}{input}\SetKwInOut{Output}{output}
  \small
  \vspace{1mm}
  \textbf{Input:} Randomized Images $\xb^{\rm ran}$, Canonical Images $\xb^{\rm can}$, Sensor Informations $\bb$, States $\zb$, Control Inputs $\ub$ \\
  Initialize encoder network $\phi$, decoder network $\theta$, dynamics parameters $\gamma$, DPN $\psi$, pseudo initial observation $\yb_0$\\
      $\alphab_1 \leftarrow \psi(\yb_{0}, \ub_{1})$ \\
      $\gamma_1 \leftarrow \gamma_{1}(\alphab_1)$ \\
      Compute $\log p(\zb_1)$ \\
      \For{$step ~ t = 1, 2, ...,T$}
      { 
        Sample $\tilde{\ab}_t \sim q_{\phi}(\ab_t | \xb^{\rm ran}_t)$ \\
        Compute $\log q_{\phi}(\tilde{\ab}_t |\xb^{\rm ran}_t)$  \Comment{\textcolor{magenta}{Use Randomized Image}}\\
        Compute $\log p_{\theta}(\xb^{\textrm{can}}_t|\tilde{\ab}_t$)  \Comment{\textcolor{magenta}{Use Canonical Image}}\\
        $\tilde{\yb}_t \leftarrow [\tilde{\ab}_t, \bb_t] $ \\
        Compute $\log p_{\gamma_t}(\tilde{\yb}_t|\zb_t)$ \\
        Predict $\alphab_{t+1} \leftarrow \psi(\tilde{\yb}_{0:t}, \ub_{1:t+1})$ \\
        $\gamma_{t+1} \leftarrow \gamma_{t+1}(\alphab_{t+1})$ \\
        Compute $\log p_{\gamma_{t+1}}(\zb_{t+1}|\zb_{t}, \ub_{t+1})$ \\
      }
      Compute ELBO \eqref{eq:KRCAN_ELBO_clear} using Monte Carlo integration with samples with $\tilde{\ab}$\\
      Update model parameters $\theta, \gamma, \psi, \phi$, using stochastic gradient ascent
\caption{KRC-model Training (1sequence)}
\label{alg:KRC-model}
\end{algorithm}

\begin{algorithm}[t]
  \SetKwData{Left}{left}\SetKwData{This}{this}\SetKwData{Up}{up}
  \SetKwFunction{Union}{Union}\SetKwFunction{FindCompress}{FindCompress}
  \SetKwInOut{Input}{input}\SetKwInOut{Output}{output}
  \textbf{Input:} KRC-model, Cost Function $C$, Task Horizon $L$, Planning Horizon $H$, CEM Parameters \\ 
    \For{$step ~ l = 1, 2, ...,L$}
    { 
    Observe $\xb^{\textrm{real}}_l, \bb_l$\\
    Sample $\tilde{\ab}_l \sim q_{\phi}(\ab_l | \xb^\textrm{real}_l)$ \Comment{\textcolor{magenta}{Use Real Image}}\\
    $\hat{{\zb}}_l \leftarrow \textrm{Kalman Filtering}(\tilde{\ab}_l,  \bb_l)$ \\
    Optimize ($\ub^*_{l+1}, \dots \ub^*_{l+H}) \leftarrow \text{CEM}(C(\cdot))$ \\
    Execute $\ub^*_{l+1}$ \\
    }
\caption{KRC-MPC}
\label{alg:KRC-MPC}
\end{algorithm}

    \subsection{KRC-MPC} \label{KRC-MPC}
        By using learned KRC-model as a visual dynamics model, zero-shot sim-to-real transferable visual MPC becomes possible. 
        We call this framework KRC-MPC.
        In this framework, the intrinsic feature $\ab_t$ is extracted from the real image $\xb^\textrm{real}_t$ by RCAN's encoder $q_{\phi}(\ab_t|\xb_t^\textrm{ran})$, and then the state of the dynamics is estimated by the Kalman Filter of the dynamics model, using it at each time step.
        Subsequently, a control input sequence is optimized by planning, as formulated in Eq.~\eqref{planning}, so that the cost function between the predicted states by the model and the target states is minimized: 
        \begin{equation} \label{planning}
        \left.
        \begin{alignedat}{2}
            &\ub^*_{l+1}, \dots \ub^*_{l+H} && = \argmin_{\ub_{l+1}, \dots \ub_{l+H}}  \frac{1}{H} \sum_{\tau=l+1}^{l+H}  C(\sb_\tau) \\
            &\text{subject to} ~~~ \sb_\tau && = p_{\gamma_\tau, \psi}({\bf z}_\tau \mid {\bf z}_{\tau-1}, {\bf u}_\tau) .
        \end{alignedat}
        \right\}
        \end{equation}
        Here, $H$ is the planning horizon, $\tau$ is the planning index, $C$ is the cost function, and
        $\ub^*_{l+1}, \dots \ub^*_{l+H}$ is the optimal control input sequence.
        The process of KRC-MPC using KRC-model is shown in Algorithm \ref{alg:KRC-MPC}.

\section{Simulation} \label{Simulation}
    To verify the effectiveness of our method, i.e., KRC-model learning and KRC-MPC, we applied it to a valve rotation task and a block mating task. These tasks are suitable for evaluating our method because it is domain-independent; only the object's and robot's states are required rather than detailed domain information, such as the object's color, lighting condition, background, etc., for its task execution. Thus, we utilize canonical images where the object's and robot's states can be identified, as shown in Fig.~\ref{fig:overview_of_mehod}. 

    Our simulation experiments can be summarized as follows:
    \begin{enumerate}
        \item Evaluation of control performance in various appearances of simulation domains.
        \item Ablation study for canonical images
    \end{enumerate}

    \subsection{Simulation Environment}
        \subsubsection{Valve Rotation Task}
            We used the ROBEL D'Claw environment\cite{Ahn_CoRL_2019} built in the physics simulator MuJoCo\cite{mujoco} as shown in Fig.~\ref{fig:robel_simulation_env} (a). This environment consists of a 9-degrees-of-freedom (DoF) robot hand
            where each finger has 3-DoF and a 1-DoF valve. 
        \subsubsection{Block Mating Task}
\begin{figure}[t]
    \centering
    \includegraphics[width=0.95\columnwidth]{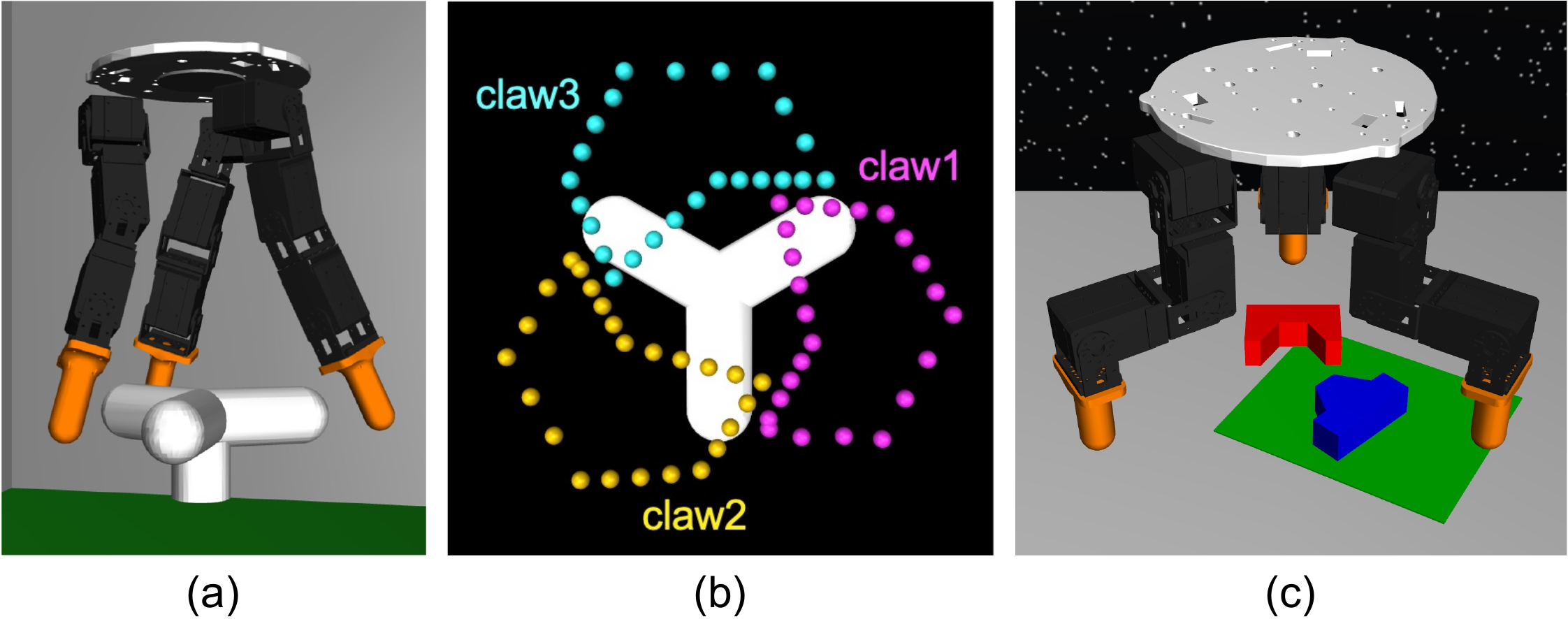}
    \caption{
        (a) Environment for valve rotation task (b) Action space of valve rotation task (c) Environment for block mating task. In (c), the green square area means the action space of block mating task (only for visualization).
    }
    \label{fig:robel_simulation_env}
    \vspace{-2mm}
\end{figure}

            We built a block mating environment \cite{Suraj_ICRA_2020} consisting of two
            blocks: one fixed female block and one free male block, as shown in Fig.~\ref{fig:robel_simulation_env} (c). The fixed female block has 1-DoF corresponding to the horizontal position while the free male block has 3-DoF including horizontal position, vertical position, and z-axis rotation. Then, we used only one finger of D'Claw as a robot.

    \subsection{Model Settings}
        \subsubsection{State Space} \label{subsubsection:state_spce}
        $<${\bf Valve Rotation Task}$>$
        For the robot, only the first and second joints are controllable in each finger, resulting in six dimensions across three fingers. For the valve, we used $[\cos{(3\theta^\textrm{valve})}, \sin{(3\theta^\textrm{valve})}]$, where $\theta^{\textrm{valve}}$ is its joint angle. Thus, the dimension of the state space is $d_\zb = 8$.
        $<${\bf Block Mating Task}$>$
        For the robot, the fingertip position is constrained to a two-dimensional plane.
        For the blocks, the total DoF is 6, but the male's z-axis rotation is expressed as $[\cos{(\theta^\textrm{block})}, \sin{(\theta^\textrm{block})}]$, thus, $d_\zb = 7$. Also, each state was normalized to $[0, 1]$ in both environments.

        \subsubsection{Action Space} \label{subsec:action_space}
            $<${\bf Valve Rotation Task}$>$
            We defined action as the position of each fingertip and constrain them to a one-dimensional (1D) manifold.
            Each colored dotted line in Fig.~\ref{fig:robel_simulation_env} (b) corresponds to the manifold, and each point on the lines represents the position.
            However, their discontinuity  between 0 and 1 is not desirable for model learning.
            Therefore, in model learning, the target positions of the controllable joints are used as the actions; So, $d_\ub = 6$. 
            Also, each action was normalized to $[0, 1]$.
            $<${\bf Block Mating Task}$>$ The fingertip positions on a 2D plane in which each axis is normalized to $[0, 1]$ are used as an action space in all of model learning, data collection and MPC. So, $d_\ub = 2$. 
        
        \subsubsection{Observation Space}
            $<${\bf Valve Rotation Task}$>$
            We used a $64 \times 64$ RGB image and the encoder value of all finger joints (6D in total) as observations.
            Then, we set the dimension of the intrinsic feature to $d_\ab = 8$. 
            Thus, the total dimension of observation for dynamics model is $d_\yb = 14$, including both the intrinsic feature and encoder values of all finger joints.
            $<${\bf Block Mating Task}$>$
            We used a $64 \times 64$ RGB image and the 2-DoF fingertip position as observations. 
            Thus, we set $d_\ab = 8$ and $d_\yb = 10$.
             The detailed settings are on our project page\footnote{ https://tomoya-yamanokuchi.github.io/krcmpc/}.

\begin{table}[t]
    \caption{
            Randomized parameters and their ranges and applied intervals.
    }
    \label{tab:randomized_parameters}
    \vspace{-2mm}
    \begin{center}
        {\scriptsize
            \begin{tabular}{l|l|l}
                \toprule
                \textbf{Randomized Parameter} & \textbf{Parameter Distribution}  & \textbf{Interval} \\ 
                \midrule
                texture                     & uniform ([0, 255])         &  each step/sequence \\
                camera position (x-axis)    & uniform ([-0.006, 0.006])  &  each sequence \\
                camera position (y-axis)    & uniform ([-0.006, 0.006])  &  each sequence \\
                camera position (z-axis)    & uniform ([0.31, 0.34])     &  each sequence \\
                light position              & uniform ([0, 16])          &  each sequence \\
                \bottomrule
            \end{tabular}
        }
    \end{center}
    \vspace{-1.5mm}
\end{table}

    \subsection{Training Data Collection} \label{Collection_of_training_dataset}
        \subsubsection{Valve Rotation Task} \label{datacollection:valve_rotation}
            We collected 1500 sequences of self-supervised data generated by random actions. 
            We additionally collected one sequence of task-specific data such that the robot rotates a valve at a constant velocity of 0.157 [rad/sec], and 500 sequences generated by applying Gaussian noise to the task-specific sequence, in order to simplify the data collection.
             Thus, the total sequences are $N=2001$.
            Visual randomization was applied when collecting data.
            The randomized parameters, ranges and applied intervals are shown in Table~\ref{tab:randomized_parameters}.
            We applied control input sequences and collected the pair of randomized image, canonical image, encoder value, state, and control input at each time step. 
            The randomized and canonical image for extracting the intrinsic features were jointly rendered at a one simulation instance.
            
        \subsubsection{Block Mating Task}
            We collected 2005 sequences for model learning. The sequences consist of five task-specific sequences such that the robot mates the block pair with five different horizontal positions of female block (-3, -1.5, 0, 1.5, 3 [cm]), and 2000 sequences generated by applying Gaussian noise to the task-specific sequences.
            The same visual randomization as described in IV.C.1) was applied when collecting the data, except for the range of the camera position (z-axis), which was changed to uniform ([0.185, 0.198]). Also, the length of each sequence is $T=25$ in both the tasks.

    \subsection{Results} \label{subsection:result_simulation}
        \subsubsection{Evaluation of control performance in various appearances of simulation domains}
            To evaluate the effectiveness of our visual MPC, we compared KRC, KR2, KC2, and Random Policy. 
            Additionally, in order to evaluate the difference in control performance depending on the observability of the state $\zb$, KRC and KR2 were evaluated in both the cases where the state $\zb$ was observable and unobservable. 
            We randomized the object's color and texture, lighting conditions, and camera positions in the test domains, as was done during training data collection, to investigate zero-shot transferability of the proposed method to such new domains. 
            Basically, visual randomization was applied per each sequence of data, but exceptional treatments were made depending on the methods. 
            
            The details of these methods are as follows: 
            \begin{itemize}
                \item KRC {\it w/} $\zb$: Proposed visual MPC with observed state. Texture randomization is applied per each step. 
                \item KRC {\it w/o} $\zb$: Proposed visual MPC with unobserved state. Texture randomization is applied per each step. 
                \item KR2 {\it w/} $\zb$: MPC which uses a model whose input and output images are both common randomized images with observed state. Texture randomization is applied per each step. 
                \item KR2 {\it w/o} $\zb$: MPC which uses a model whose input and output images are both common randomized images with unobserved state. 
                Texture randomization is applied per each sequence, unlike KR2 {\it w/} $\zb$. This corresponds to the framework of previous works \cite{Ryan_RSS_2020,Suraj_ICRA_2020}.
                \item KC2 {\it w/} $\zb$: MPC which uses a model whose input and output images are both canonical images. This is the ablative variant of the proposed method in which no visual randomization is applied to the input images. 
                \item Random Policy: The policy that execute the action randomly sampled. This is the baseline to evaluate the task difficulty.
            \end{itemize}
            The detailed MPC settings are on our project page$^{1}$

            $<${\bf Valve Rotation Task}$>$
            We applied methods described above to a valve rotation task, in which the valve is rotated at a constant velocity of 0.157 [rad/sec], and then evaluated the control success rate on 100 test domains.
            We counted a task as successful if the value of the evaluation function
            \begin{align}
                J = \frac{1}{L}\sum_{l=1}^{L} | \theta^{\textrm{valve}*}_l - \theta^{\textrm{valve}}_l |
                \label{eq: task_eval} 
            \end{align}
            was lower than the baseline, where $\theta^{\textrm{valve}*}_l$ and $\theta^\textrm{valve}_l$ are the target valve position and the actual valve position at time step $l$, respectively. 
            We defined the baseline as the value coming from the execution when the control inputs of the task-specific sequence are used for control in an open-loop manner.
            The task execution steps was set to $L=20$. 

            The Cross-Entropy Method (CEM) \cite{Reuven_MCAP_1999_CEM} was used for task execution by MPC.
            We used the cost function
            \begin{align} \label{eq: cem_cost}
            C(\sb_\tau) = \underbrace{\| \zb^{\textrm{object}*}_{\tau} - \hat{\zb}^\textrm{object}_{\tau} \|^2}_{\text{tracking cost}} + \beta \cdot \underbrace{\mathbb{V}_{\omega}[f_\omega (\hat{\zb}_{\tau-1}, \ub_\tau)]}_{\text{variance cost}}
            \end{align}
            for CEM, where $\hat{\zb}_{\tau}$ is the state predicted by the model based on Eq.~\eqref{eq: LGSSM-transition}, $\hat{\zb}^\textrm{object}_{\tau}$ is the state of only the dimension related to the objects (in this task, $\hat{\zb}^\textrm{object}_{\tau}$ = $\hat{\zb}^\textrm{valve}_{\tau}$) in $\hat{\zb}_{\tau}$, and $\zb^\textrm{object*}_{\tau}$ is the target state corresponding to $\hat{\zb}^\textrm{object}_{\tau}$.
            In Eq.~\eqref{eq: cem_cost}, the first term is the cost for tracking the target valve trajectory, the second term is the variance cost by the ensemble network \cite{cem2} to avoid uncertain actions, and $\beta$ means the weight of the variance cost.  We tuned $\beta$ to roughly match the scales of tracking cost and variance cost.
            The ensemble network consists of $M$ independent networks $\{f_{\omega_1}, \dots, f_{\omega_M}\}$, each of which is trained as a dynamics model in state space, separately from KRC-model.

\begin{table}[t]
    \vspace{1mm}
    \caption{
            Control success rates for the valve rotation task and control errors for the block mating task
            in various appearances of simulation domains. For each method, models were pre-trained with five different seeds, and 100 trials were performed by randomly sampling one of these five models for each task execution.
            \label{tab:result_visual_ablation_simulation}
    }
    \vspace{-2mm}
        \begin{center}
         \footnotesize{
            \begin{tabular}[b]{l|c|c}
                \toprule
                \begin{tabular}[b]{c} \vspace{0.8mm} \textbf{Model} \\  \end{tabular} & \begin{tabular}[b]{c} \textbf{Valve Rotation} \\ \textbf{(Success)} \end{tabular}  & \begin{tabular}[b]{c} \textbf{Block Mating} \\ \textbf{(Error)} \end{tabular} \\ 
                \midrule
                KRC {\it w/} {\bf z}        & \textbf{99/100}   & ${\bf 0.897\pm{0.538}}$ \\
                KRC {\it w/o} {\bf z}       & \textbf{100/100}  & ${\bf 0.436\pm{0.514}}$ \\
                KR2 {\it w/} {\bf z}        & 95/100            & $1.305\pm{0.609}$ \\ 
                KR2 {\it w/o} {\bf z}       & 74/100            & $2.359\pm{1.075}$ \\ 
                KC2 {\it w/} {\bf z}        & 2/100             & $3.856\pm{2.070}$ \\ 
                Random Policy               & 0/100             & $2.595\pm{1.827}$ \\  
                \bottomrule
            \end{tabular}
        }
        \end{center}
\end{table}

\begin{figure}[t]
\centering
    \begin{minipage}[b]{\columnwidth}
        \centering
        \includegraphics[width=0.5\columnwidth]{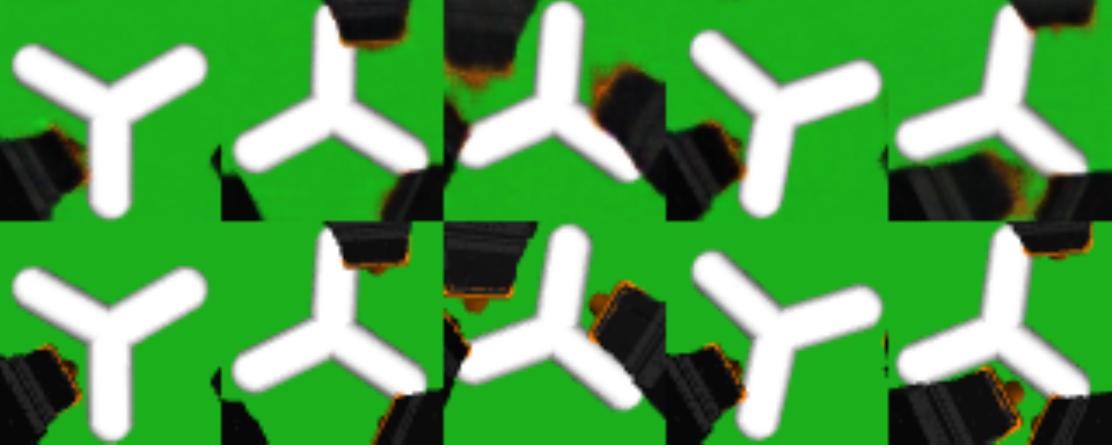}
        \caption{
            ({\bf Upper}) A qualitative result on predictions of the dynamics model of KRC {\it w/} $\zb$, except for the image on the left end which corresponds to initial observed step ({\bf Lower}) The true images corresponding to the predicted images
        }
        \label{fig:images_model_predict_and_true}
        \vspace{3mm}
    \end{minipage}
    \begin{minipage}[b]{\columnwidth}
        \centering
        \includegraphics[width=1\columnwidth]{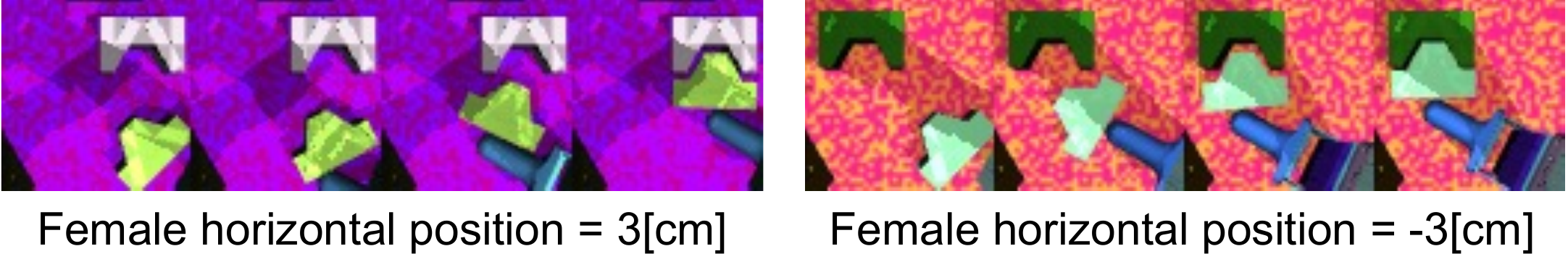}
        \caption{
            Examples of a snapshot during task execution of block mating task by KRC {\it w/o} $\zb$ in simulation
        }
        \label{fig:snapshot_block_mating}
        \vspace{-2mm}
    \end{minipage}
\end{figure}

            The results of control success rates are shown in Table~\ref{tab:result_visual_ablation_simulation}.
            We can see that KRC achieved better performance than the other methods. Then, the results show that even when the state is unobservable, the control performance is comparative to that of the case where the state is observable. We show a qualitative result on predictions of the dynamics model of KRC-model in Fig.~\ref{fig:images_model_predict_and_true}. From Fig.~\ref{fig:images_model_predict_and_true}, we can see that the KRC-model can predict future states precisely.   
        
            $<${\bf Block Mating Task}$>$
            We applied the methods to the block mating task and evaluated the error between the target male block state $\zb^{\rm block*}$ and the manipulated male block state $\zb^{\rm block}$. The error is defined as $\frac{1}{L}\sum_{l=1}^{L} \| \zb^{\rm block*}_l - \zb^{\rm block}_l \|$. The test domains consist of five different block positions, each of which includes 20 test cases, resulting in 100 test cases with the different visual domains. The task execution step's horizon was set to $L=20$. For the CEM cost function, we used same cost function in Eq. (18) with block state $\zb^{\rm block}$. 
            The results are shown in Table~\ref{tab:result_visual_ablation_simulation}. We can see that KRC achieved better performance than the other methods. In addition, KRC {\it w/o} {\bf z} performed better than KRC {\it w/} {\bf z}. 
            This is probably due to the inconsistency between image observations and states, caused by the visual randomization. Namely, since the perception of block positions is inherently considered to change depending on the camera positions; therefore, if camera positions are randomized, the supervised data of block positions should be relative to the camera positions, whereas the absolute values were given in this experiments (Conversely, since the rotation angle is not sensitive to changes in camera positions, the performance of KRC {\it w} {\bf z} and KRC {\it w/o} {\bf z} may be comparative for the valve rotation task). The snapshot during task execution by  KRC {\it w/o} {\bf z} is shown in Fig.~\ref{fig:snapshot_block_mating}. 
            
            In summary, all of the simulation results show that KRC-MPC can be transferred to various appearances of test domains in a zero-shot manner.

\begin{figure}[t]
    \centering
    \includegraphics[width=0.9\columnwidth]{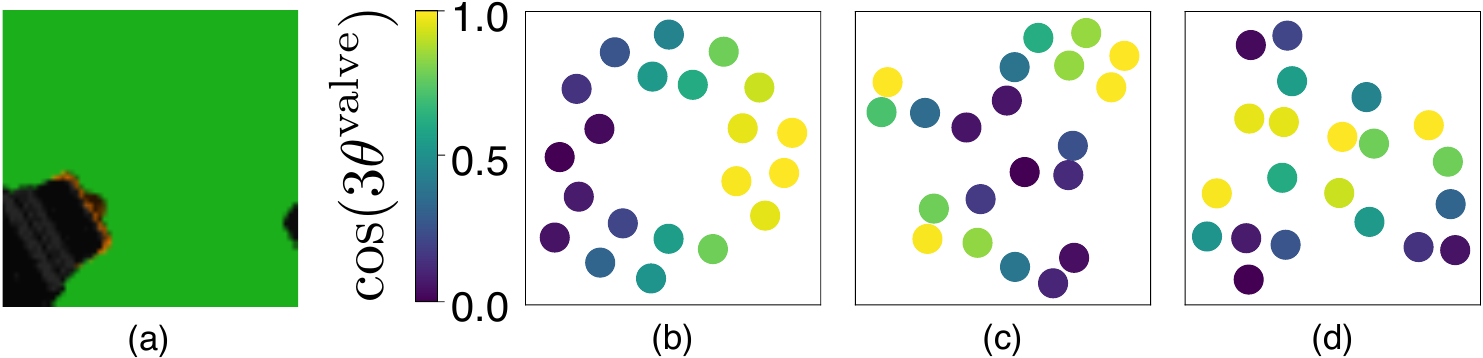}
    \caption{
    ({\bf a}) Non-canonical image with a transparent valve.
    ({\bf b}), ({\bf c}) and ({\bf d}) are the visualization of 2D latent space of intrinsic feature for KRC, KRNc with randomized images, and KRNc with transparent valve respectively. 
    The color of each point shows the value for valve angle ($\cos(3\theta^{\textrm{valve}})$). 
    }
    \label{fig:latent_space}
\end{figure}

\begin{figure}[t]
\centering
    \begin{minipage}[b]{0.24\columnwidth}
        \vspace{-2mm}
        \centering
        \includegraphics[width=0.65\linewidth]{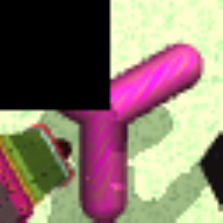}
        \caption{
            Example of image with occlusion
            \label{fig:domain_with_occlusion}
        }
    \end{minipage}
    \begin{minipage}[b]{0.68\columnwidth}
        \centering
            \captionof{table}{
                Control success rates in challenging conditions.
                For each method, models were pre-trained with five different seeds, and 100 trials were performed by randomly sampling one of these five models for each task execution.
                \label{tab:success_rate_in_occluded_images}
            }
            \footnotesize{
                \begin{tabular}[b]{l|c}
                    \toprule
                    \hspace{23mm}\textbf{Model}       & \textbf{Success}  \\ 
                    \midrule
                    KRC                               & \textbf{93/100}   \\
                    KRNc with randomized image        & 65/100            \\ 
                    KRNc with transparent valve & 34/100            \\ 
                    \bottomrule
                \end{tabular}
            }
    \end{minipage}
\end{figure}

        \subsubsection{Ablation study for canonical images}
            To understand the influence of canonical images on the learning of the latent space of intrinsic features, we visualized the latent space of models trained by using canonical and non-canonical images. We call the model with non-canonical images as KRNc. Here, for the non-canonical images, it is assumed that task-relevant objects are visually unidentifiable, unlike canonical images. In this experiment, we defined the two types of non-canonical images: (1) randomized images (2) images with a transparent valve, as shown in Fig.~\ref{fig:latent_space} (a).
            
            We first obtained intrinsic features $\ab \in \Rbb^8$ by inputting image sequences of a successful task execution of simulation experiment 1) into the feature extractor $q(\ab | \cdot)$ of each model, and then we plotted them in 2D space by using t-SNE. The visualization results are shown in Fig.~\ref{fig:latent_space} (b)-(d), in which each point is colored with respect to its corresponding valve state. We can see that the latent space of KRC is learned to maintain the similarity of valve states, which is a dominant object in images, while that of the two types of KRNc does not show such a property.
            
            To evaluate the impact of such differences in the latent space of intrinsic features on the robustness of the control performance, we compared control success rates of these three methods in visual MPC settings under challenging conditions. Specifically, we performed the valve rotation task with occluded observation images, under the initial state $\zb_1$ applied with Gaussian noise $\mathcal{N}(0, 100)$, in 100 test domains. An example of images with occlusion is shown in Fig.~\ref{fig:domain_with_occlusion}. The results of the control success rates are shown in Table~\ref{tab:success_rate_in_occluded_images}. We can confirm that the control success rates of KRC are higher than those of KRNc. This indicates that appropriate canonical images are crucial for learning a latent space of intrinsic features providing robust control.

\section{Real Experiments} \label{Real_Robot_Experiments}
    To evaluate the effectiveness of the proposed method in the real world, we applied KRC-MPC to valve rotation tasks and conducted the following two visual MPC experiments:
    \begin{enumerate}
        \item Evaluation of control success rates in various appearances of real domains.
        \item Evaluation of control performance in additional tasks.
    \end{enumerate}

\begin{figure}[t]
\centering
    \begin{minipage}[b]{\columnwidth}
        \centering
        \includegraphics[width=\columnwidth]{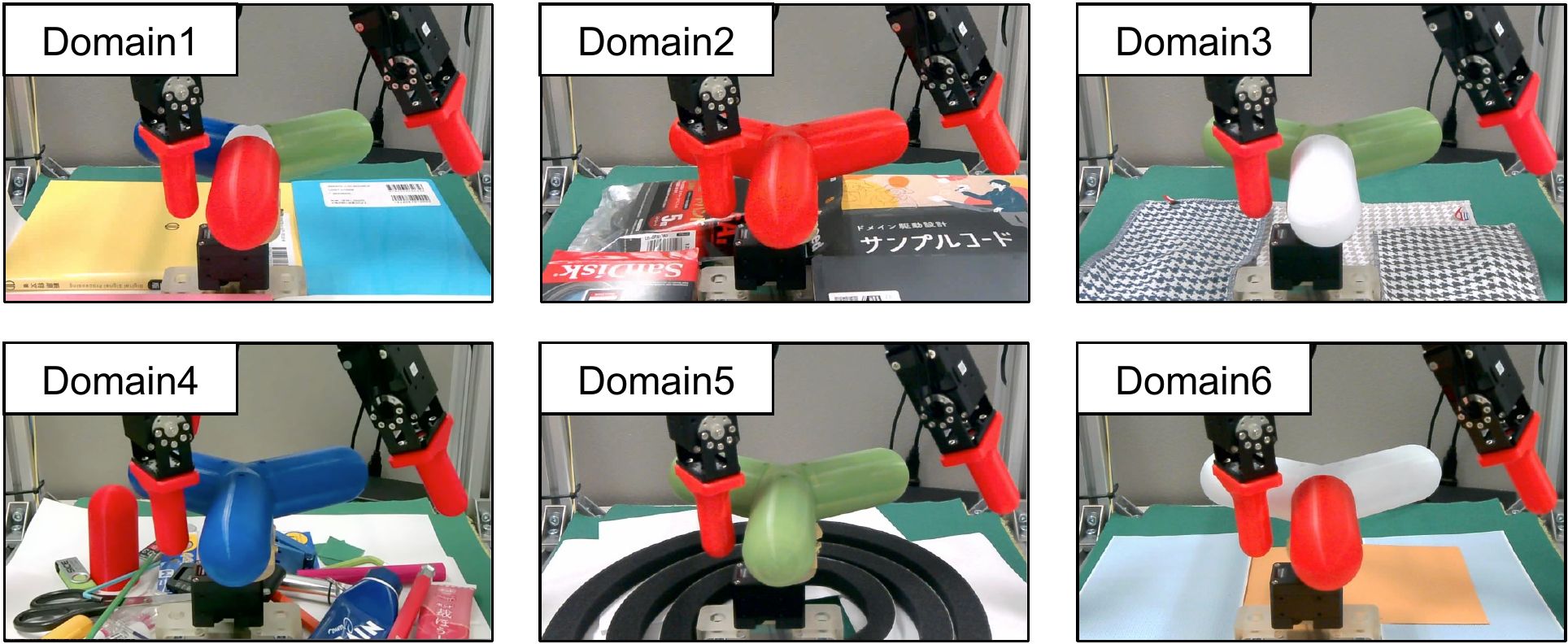}
        \caption{
            Real test domains
        }
        \label{fig:real_test_domains}
        \vspace{2mm}
    \end{minipage}
    \begin{minipage}[b]{\columnwidth}
            \centering
            \captionof{table}{
                  Control success rates in real test domains
                  \label{tab:result_real_task}
                  \vspace{-2mm}
            }
            \footnotesize{
                \begin{center}
                    \begin{tabular}{l|rrrrrr}
                        \toprule
                        \textbf{Test Domain}  &    1             &       2         &         3        &         4        &         5        &    6  \\ 
                        \midrule
                        KRC {\it w/}  {\bf z} &  \textbf{10/10}  &  \textbf{10/10} &   9/10           &    9/10    &  \textbf{10/10}  &   8/10 \\
                        KRC {\it w/o} {\bf z} &  \textbf{10/10}  &  \textbf{10/10} &   8/10           &    \textbf{10/10}   &   9/10           &   \textbf{9/10} \\
                        KR2 {\it w/}  {\bf z} &  \textbf{10/10}  &   7/10          &   7/10           &    8/10    &   3/10           &   8/10 \\
                        KR2 {\it w/o} {\bf z} &   8/10           &   8/10          &  \textbf{10/10}  &    7/10    &   9/10           &   \textbf{9/10}  \\
                        KC2 {\it w/}  {\bf z} &   0/10           &   0/10          &   0/10           &    0/10    &   0/10           &   0/10  \\
                        Random Policy         &   0/10           &   0/10          &   0/10           &    0/10  &   0/10           &   0/10 \\
                        \bottomrule
                    \end{tabular}
                \end{center}
            }
    \end{minipage}
\end{figure}

    \subsection{Results}
        \subsubsection{Evaluation of control success rates in various appearances of real domains}
\begin{figure}[t]
    \centering
    \begin{subfigure}[h]{\linewidth}
        \centering
        \includegraphics[width=0.925\linewidth]{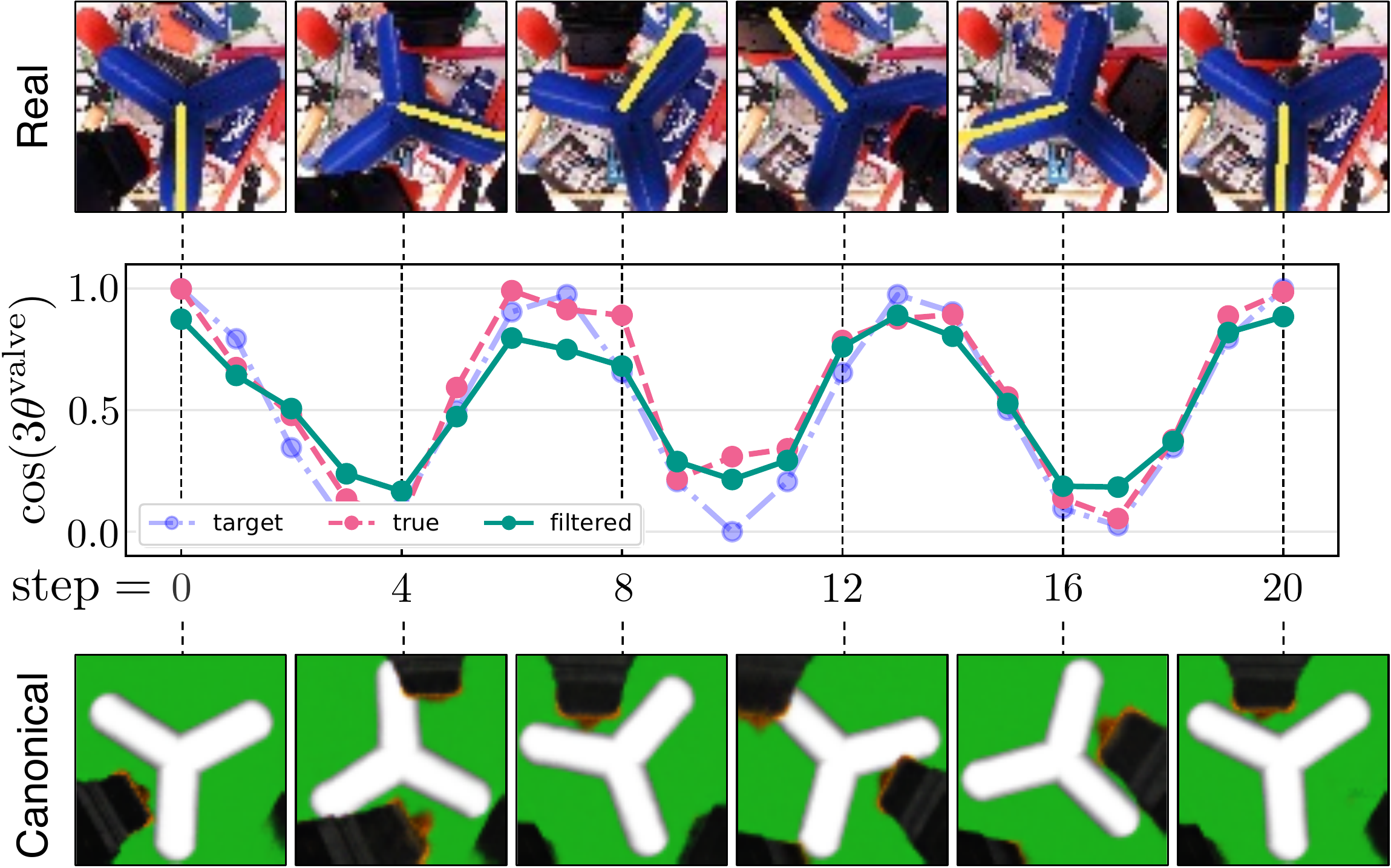}
        \caption{KRC  {\it w/} {\bf z}}
        \label{fig:snapshot_detail_Randomize_new}
        \vspace{2mm}
    \end{subfigure}
    \hspace{3.5mm}
    \begin{subfigure}[h]{\linewidth}
        \centering
        \includegraphics[width=0.925\linewidth]{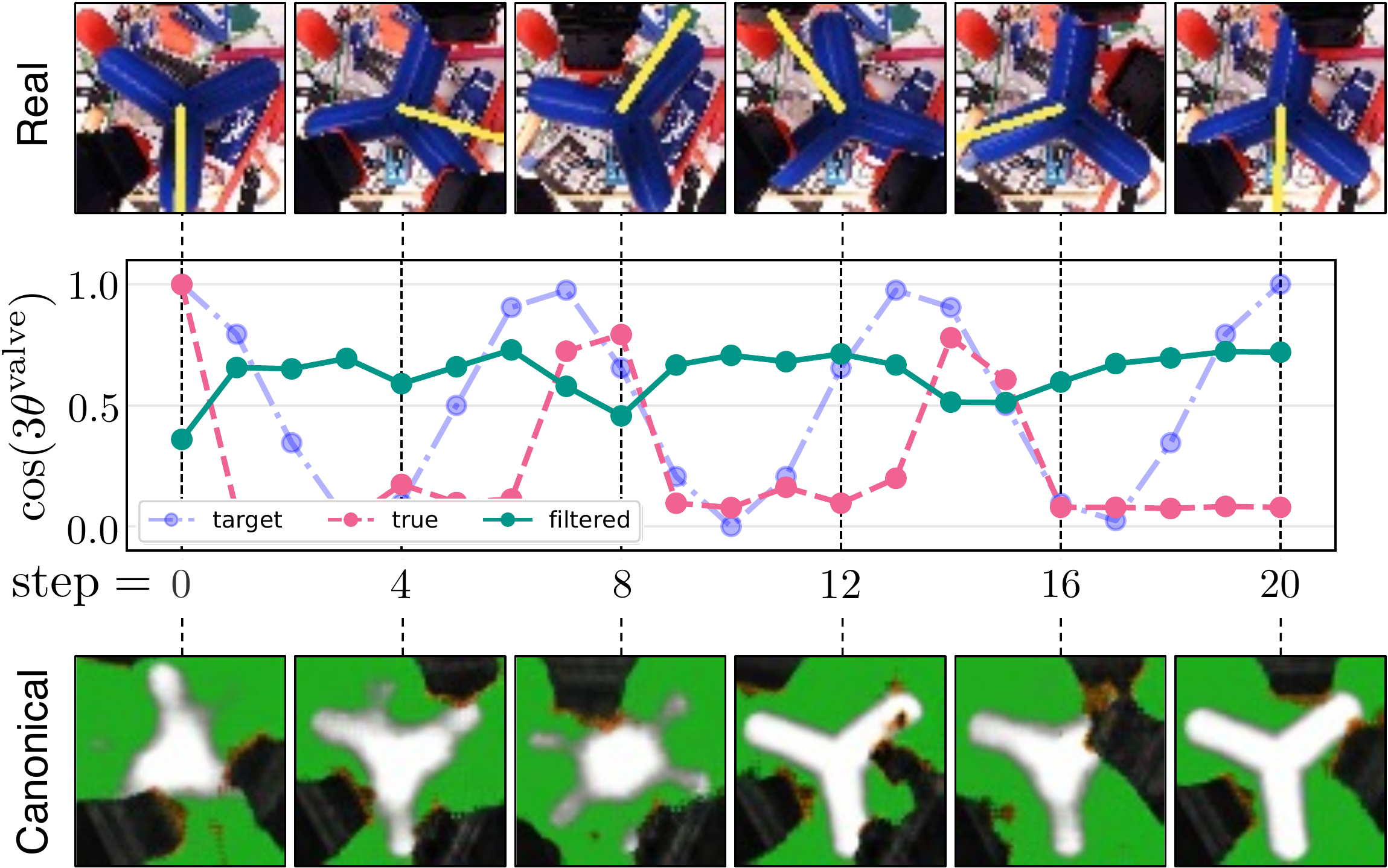}
        \caption{KC2 {\it w/} {\bf z}}
        \label{fig:snapshot_detail_No_Randomize_new}
    \end{subfigure}
    \caption{
        Example of the real images ({\bf Upper}), filtered states ({\bf Center}), and canonical images ({\bf Lower}) during task execution in Domain4.
        In each real image, the target valve angle is drawn for understanding (yellow line). 
    }
    \label{fig:snapshot_detail}
\end{figure}

            We compared control success rates of all methods in six real domains as shown in Fig.~\ref{fig:real_test_domains}. The results are shown in Table~\ref{tab:result_real_task}. We could confirm that the control success rates of KRC are better than the other methods in almost real test domains, while KC2 and Random Policy did not succeed in the task even once there. As a more detailed comparison result, we shows the estimated states and the canonical images reconstructed from them during the task execution in Domain4 for KRC  {\it w/} {\bf z} and KC2  {\it w/} {\bf z} in Fig.~\ref{fig:snapshot_detail}.
            We can see that the estimated states of KC2  {\it w/} {\bf z} are inaccurate, and the geometric configurations of canonical images are also significantly different from real images. In contrast, KRC  {\it w/} {\bf z} provides more accurate state estimation from the real images, and thus the canonical images also show accurate recognition of the real world. From these results, we found that KRC-MPC using KRC-model can be transferred to various appearances of real domains in a zero-shot manner.

        \subsubsection{Evaluation of control performance in additional tasks}
            To evaluate the transferability of our model to various tasks other than the initial task, i.e., rotation of the valve at a constant velocity of 0.157 [rad/sec], we applied KRC  {\it w/} {\bf z} and KC2  {\it w/} {\bf z} to four new tasks.
            These additional tasks were designed so that the target speed of valve rotation would change from the initial task.
            Specifically, we considered four valve rotation speeds: 0.75x, 0.5x, 0.25x, and 0.0x.
            This evaluation is conducted in Domain1.
            Table~\ref{tab:task_ablation} shows the results of the control performance. 
            Unlike the initial task, these additional tasks are evaluated by control performance based on Eq.~\eqref{eq: task_eval} due to lack of baselines for successful evaluation. We can see that KRC  {\it w/} {\bf z} has better control performance than KC2  {\it w/} {\bf z}.

\begin{figure}[t]
\centering
    \begin{minipage}[b]{\columnwidth}
        \centering
        \captionof{table}{
              Control performances in additional tasks.
              The mean and standard deviation of the three results with the lowest values of Eq.~\eqref{eq: task_eval} among the five models are shown to eliminate the outlier values for failure cases.
              \label{tab:task_ablation}
        }
        \footnotesize{
          \begin{tabular}{c|cc}
            \toprule
            \textbf{Task} & KRC {\it w/} {\bf z}  & KC2 {\it w/} {\bf z} \\
            \midrule
            0.75x  & $\mathbf{0.162 \pm{0.013}}$  &  $0.553 \pm{0.128}$ \\
            0.50x  & $\mathbf{0.127 \pm{0.012}}$  &  $0.488 \pm{0.079}$ \\
            0.25x  & $\mathbf{0.139 \pm{0.019}}$  &  $1.635 \pm{1.013}$ \\
            0.00x  & $\mathbf{0.098 \pm{0.021}}$  &  $1.183 \pm{0.470}$ \\
            \bottomrule
          \end{tabular}
        }
    \end{minipage}
\end{figure}

            In summary, all of the experimental results show that KRC-MPC can be transferred to various appearances of real domains and various tasks in a zero-shot manner.

\section{Discussion} \label{Discussion}
    Here we discuss the limitations of our method. 
    The first limitation is that our current method can only be applied to domain-independent tasks that require geometric features as task-relevant features. Thus, the domain-dependent tasks in which the target states depend on the color, texture, etc., are not covered by our current method.
    A direction of future work is to introduce an additional disentangled structure to take into account such domain-dependent information \cite{Li_ICML_2018}. Another interesting future work would be enhancing the generalization capability of the canonical feature extraction in our method. To this end, the use of contrastive learning methods may be considered \cite{Ting_ICML2020}.
    The second limitation is that it may not generalize to the domains whose dynamics are quite different from simulation. 
    Our method does not include such a mechanism to address such broad-ranging dynamics. Thus, we will expand our model for the dynamics reality gap, as done in previous works \cite{Murooka_CoRL_2020} in future work.

\section{Conclusion} \label{Conclusion}
    In this study, we proposed KRC-MPC as a framework for zero-shot sim-to-real transferable visual MPC.
    Then, we proposed KRC-model as a visual dynamics model to achieve KRC-MPC.
    Our method was evaluated by a valve rotation task in both simulation and the real world, and by a block mating task in simulation. 
    The experimental results show that KRC-MPC can be transferred to various appearances of real domains in a zero-shot manner.


\bibliographystyle{IEEEtran}
\bibliography{citations}

\section*{APPENDIX} \label{appendix}
    Learning an autoregressive model in image space \cite{Ryan_RSS_2020,Suraj_ICRA_2020} involves generating images to learn the dynamics. However, such computation is often intractable. On the other hand, as suggested by many previous studies \cite{KVAE, Limoyo_IEEE_Robotics_2020, Chiappa_METRON_2019}, the extracted features and dynamics from such high-dimensional data as images often lie on a low-dimensional manifold; therefore, capturing such a low-dimensional manifold would reduce the computational cost for model learning.
    
\begin{figure}[h]
    \centering
    \includegraphics[width=\columnwidth]{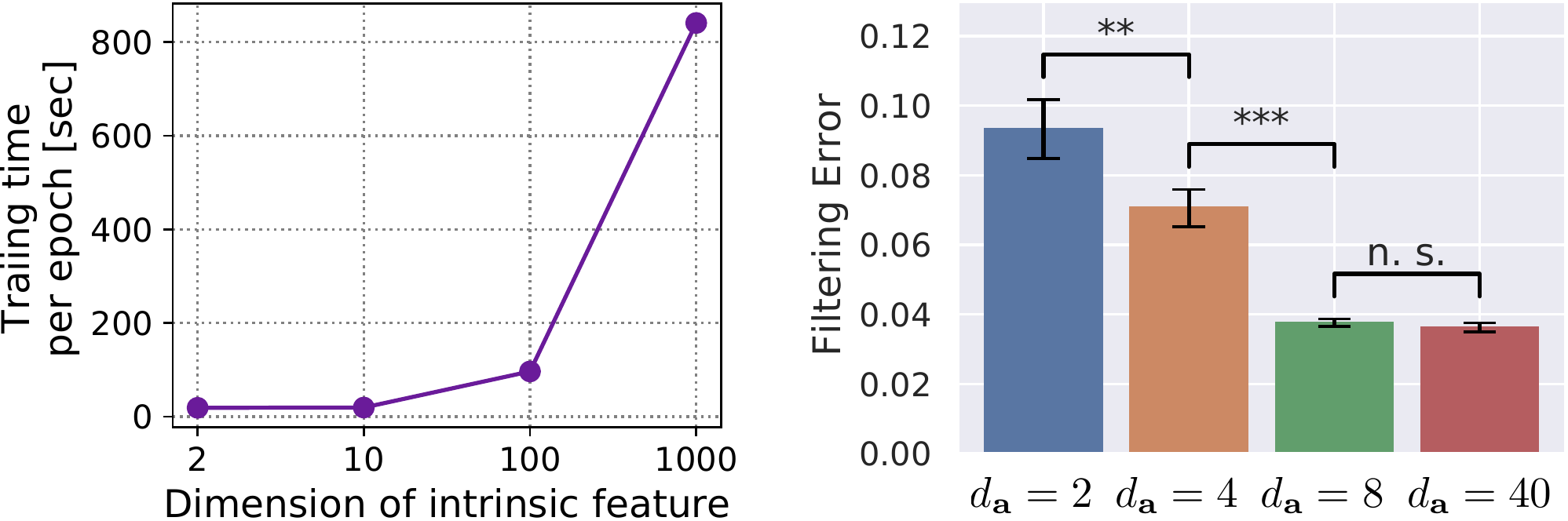}
    \caption{
    ({\bf Left}) Training time of the model per epoch (more than 4096 dimensions were intractable). Results represent the average of 10 epochs of training.
    ({\bf Right}) State estimation error for 50 sequences of off-line test data evaluated with the sum of L2-norm between the true states and estimated states by the model over all steps and sequences ($\ast$ means $p < 0.05$).
    Results represent the average of 5 models.
    }
    \label{fig:ablation_dim_a}
    \vspace{-1.7mm}
\end{figure}

    To confirm this, we evaluated training time per epoch and state estimation error for 50 sequences of off-line test data in different intrinsic feature dimensions $d_{\ab}$. Fig. \ref{fig:ablation_dim_a} shows experimental results, indicating that the training time increases rapidly as $d_{\ab}$ increases.
    We found that training was intractable in our computational environment (NVIDIA TITAN RTX, 24GB VRAM) due to running out of memory when $d_\ab \geqq 4096$. However, although the dimension $d_{\ab}$ needs to be somewhat high in terms of the state estimation error, a dimension higher than this level does not significantly contribute to decreasing the state estimation error. These results show that extracting intrinsic features from the image using a disentangled structure is crucial for maintaining model accuracy and making the learning tractable.

\end{document}

%% file: latex_settings_tomoya-y.tex
\usepackage{amsmath}
\usepackage{amssymb}
\usepackage{amsfonts}
\usepackage[font=footnotesize]{caption}
\usepackage[font=footnotesize]{subcaption}
\usepackage{booktabs} 
\usepackage[pdftex]{graphicx}
\PassOptionsToPackage{dvipsnames}{xcolor}
\usepackage{tikz}
\usepackage{siunitx}
\usepackage{multirow}
\usepackage{bbm}
\usepackage{cases}
\usepackage[ruled,vlined]{algorithm2e}
\usepackage{algorithmicx,algpseudocode}
\usepackage[maxfloats=256]{morefloats}
\usepackage{colortbl,array,xcolor}
\usepackage{mathtools}

\usetikzlibrary{positioning}
\usetikzlibrary{arrows,shapes,automata,backgrounds,fit,petri,calc}

\tikzstyle{latent} = [circle,fill=white,draw=black,inner sep=1pt,
minimum size=25pt, font=\fontsize{10}{10}\selectfont, node distance=1]

\tikzstyle{pseud} = [minimum size=25pt, font=\fontsize{10}{10}\selectfont, node distance=1]

\DeclarePairedDelimiterX{\infdivx}[2]{}{}{%
  #1\;\delimsize\|\;#2%
}

\newcommand{\argmin}{\mathop{\rm arg~min}\limits}

\colorlet{DPNColor}{orange}
\colorlet{RCANColor}{RubineRed}

\def\drm{\mathrm{d}}

\def\ub{\mathbf{u}}

\def\ab{\mathbf{a}}
\def\bb{\mathbf{b}}

\def\db{\mathbf{d}}

\def\sb{\mathbf{s}}

\def\zb{\mathbf{z}}
\def\yb{\mathbf{y}}

\def\0{\mathbf{0}}
\def\ub{\mathbf{u}}

\def\xb{\mathbf{x}}

\def\Ab{\mathbf{A}}
\def\Bb{\mathbf{B}}
\def\Cb{\mathbf{C}}

\def\Fcal{\mathcal{F}}

\def\Lcal{\mathcal{L}}

\def\Rbb{\mathbb{R}}

\def\alphab{\text{\boldmath $\alpha$}}
